\documentclass[10pt,journal]{IEEEtran}
\usepackage{graphicx}
\usepackage{subfigure}
\usepackage{amsfonts}
\usepackage{algorithm}
\usepackage{epstopdf}
\usepackage{amssymb,amsmath,amsthm}
\usepackage[breaklinks=true,bookmarks=false]{hyperref}
\usepackage[compact]{titlesec}
\usepackage{array}
\usepackage{gensymb}
\usepackage{multirow}
\usepackage{tabularx,booktabs}
\usepackage{comment}

\newcolumntype{L}[1]{>{\raggedright\let\newline\\\arraybackslash\hspace{0pt}}m{#1}}
\newcolumntype{C}[1]{>{\centering\let\newline\\\arraybackslash\hspace{0pt}}m{#1}}
\newcolumntype{R}[1]{>{\raggedleft\let\newline\\\arraybackslash\hspace{0pt}}m{#1}}
\graphicspath{{..//assets//}}

\input{tcilatex}
\begin{document}

\title{Coarse-to-Fine Multi-Scene Pose Regression with Transformers}
\author{Yoli Shavit, Ron Ferens, Yosi Keller$^{\ast}$
\IEEEcompsocitemizethanks{\IEEEcompsocthanksitem Y. Shavit, R. Ferens and Y. Keller are with the Faculty of Engineering, Bar Ilan University, Ramat-Gan, Israel.
Email: yosi.keller@gmail.com
}}
\protect \and
\maketitle

\begin{abstract}
Absolute camera pose regressors estimate the position and orientation of a
camera given the captured image alone. Typically, a convolutional backbone
with a multi-layer perceptron (MLP) head is trained using images and pose
labels to embed a single reference scene at a time. Recently, this scheme
was extended to learn multiple scenes by replacing the MLP head with a set
of fully connected layers. In this work, we propose to learn multi-scene
absolute camera pose regression with Transformers, where encoders are used
to aggregate activation maps with self-attention and decoders transform
latent features and scenes encoding into pose predictions. This allows our
model to focus on general features that are informative for localization,
while embedding multiple scenes in parallel. We extend our previous
MS-Transformer approach \cite{shavit2021learning} by introducing a
mixed classification-regression architecture that improves the localization
accuracy. Our method is evaluated on commonly benchmark indoor and
outdoor datasets and has been shown to exceed both multi-scene and
state-of-the-art single-scene absolute pose regressors. We make our
code publicly available from \href{https://github.com/yolish/c2f-ms-transformer}%
{ here}.
\end{abstract}

\section{Introduction}

\label{sec:methods}

Localizing a camera using a query image is essential for a variety of
computer vision applications, including indoor navigation, augmented
reality, and autonomous driving. Current approaches to estimating a camera's
position and orientation offer different trade-offs between accuracy,
runtime, and memory requirements. For example, hierarchical {Structure-based} localization
pipelines {(SbP)} \cite{sattler2016efficient,taira2018inloc,sarlin2019coarse}
achieve state-of-the-art (SOTA) pose accuracy but require a response time of
hundreds of milliseconds, a large memory footprint, and client-server
connectivity. These approaches employ image retrieval (IR) to find images
similar to the query image while extracting and matching local image
features. The 2D-2D matches extracted are mapped to 2D-3D correspondences
through depth or a 3D point cloud that are used to estimate the camera pose
with Perspective-n-Point (PnP) and RANSAC \cite{fischler1981random}.
\begin{figure}[tbh]
\includegraphics[width=\linewidth]{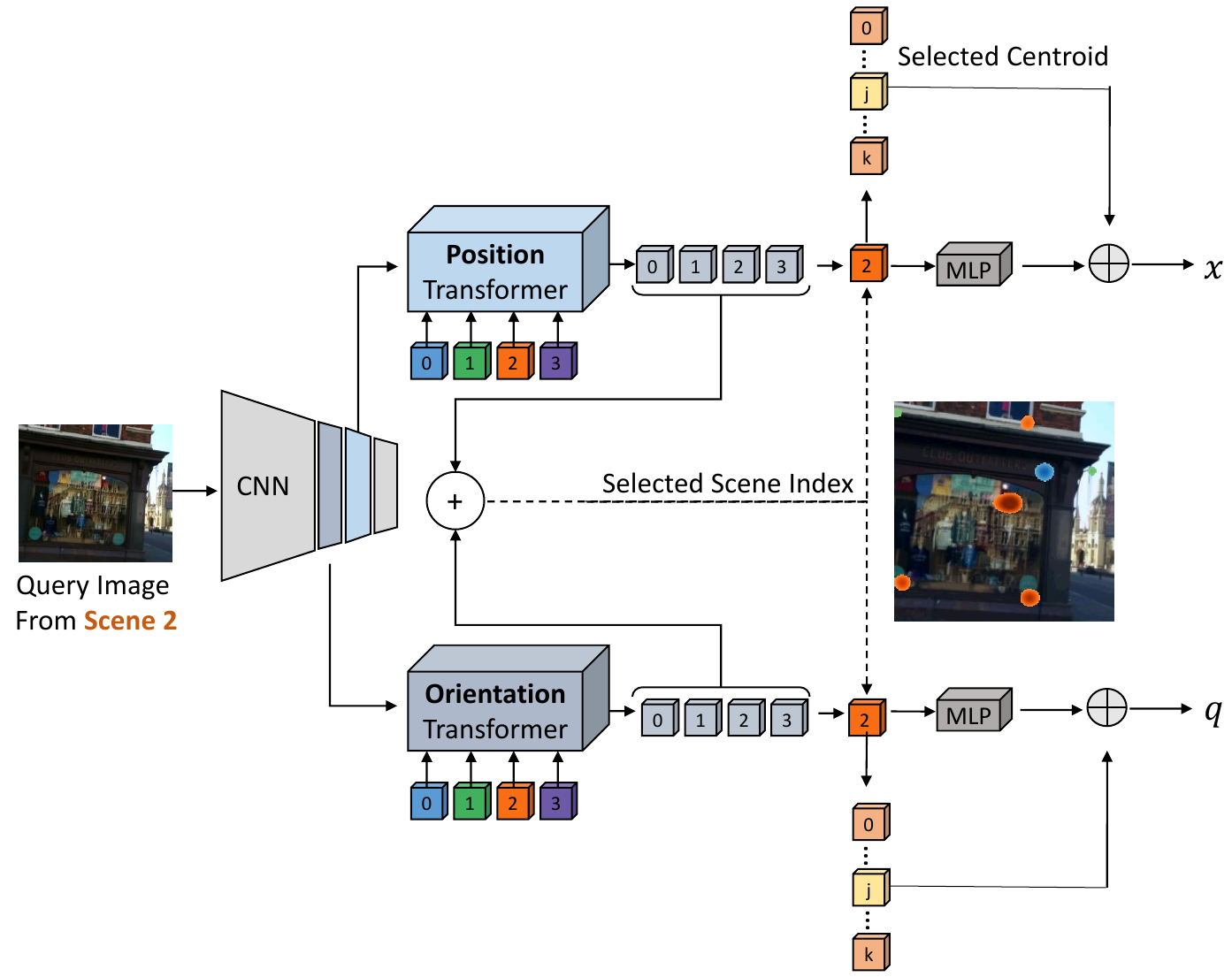}
\caption{Coarse-to-Fine Multi-scene absolute pose regression with
Transformers. Two Transformers separately attend to position- and
orientation- informative features from a convolutional backbone.
Scene-specific queries (0-3) are further encoded with aggregated activation
maps into latent representations, from which a single output is selected.
The strongest response, shown as an overlaid color-coded heatmap of
attention weights, is obtained with the output associated with the input
image's scene. The selected outputs, at the scene level, are used to further
select position and orientation centroids and regress the respective
residuals. The position $x$ and the orientation $q$ are given by the sum of
selected centroids and the regressed residuals.}
\label{fig:teaser}
\end{figure}
Absolute pose regressors (APRs), on the other hand, estimate the camera pose
in a single forward pass, using only the query image. They are faster and
can be deployed on a thin client as standalone applications due to their
small memory footprint. However, APRs are also an order of magnitude less
accurate than {SbP approaches} and other methods that use
3D data at inference time \cite{sattler2019understanding}. Furthermore, most
APRs are designed to embed a single scene at a time, implying that, for a
dataset with $N$ scenes (such as a hospital with many wards and rooms), $N$
models must be trained, deployed and selected during inference. This paper
aims at improving the accuracy of APRs by extending the current single-scene
paradigm to simultaneously learn multiple scenes. Absolute camera pose
regression was first suggested by Kendall et al. \cite{kendall2015posenet}.
Following the success of convolutional neural networks (CNNs) in multiple
computer vision tasks, the authors suggested adapting a GoogleNet
architecture to camera pose regression by attaching a multilayer perceptron
(MLP) head. A proposed architecture called PoseNet provided a fast and
lightweight solution for camera localization. However, it also suffered from
poor accuracy and limited generalization. Various absolute pose regression
methods were suggested to address these issues, proposing to modify the
backbone and MLP architecture \cite%
{naseer2017deep,walch2017image,shavitferensirpnet,wang2020atloc,cai2019hybrid}%
, as well as different loss formulations and optimization strategies \cite%
{kendall2016modelling,kendall2017geometric, shavit2019introduction}. APRs
share two common traits: (1) using a CNN backbone to produce a global latent
vector to regress the pose, and (2) training a model for each scene
(scene-specific APRs). Blanton et al. \cite{blanton2020extending} extended
single-scene absolute pose regression to a multi-scene paradigm. Similarly
to existing APRs, this method applies a CNN backbone to generate a latent
global descriptor of the image. However, instead of using a single
scene-specific MLP, it trains a set of Fully Connected (FC) layers, with a
layer per scene, which is chosen based on the predicted scene identifier.
While offering a new general framework for optimizing a single model for
multiple scenes, this method was unable to match the accuracy of
contemporary SOTA APRs.

In this work, we propose a novel formulation of multi-scene absolute pose
regression, inspired by the recent successful applications of Transformers
to computer vision tasks such as object detection \cite{DETR} and image
recognition \cite{16x16}. These works demonstrated the effectivity of
\textit{encoders} in focusing on latent features (in image patches or
activation maps) that are informative for particular tasks, by
self-attention aggregation. In addition, \textit{decoders} were shown to
successfully generate multiple independent predictions, corresponding to
queries, based on the input embedding \cite{DETR}. Similarly, we propose to
apply Transformers to multiscene absolute pose regression, using \textit{%
encoders} to focus on pose-informative features and \textit{decoders} to
transform encoded scene identifiers into latent pose representations (Fig. %
\ref{fig:teaser}). As pose estimation involves two different tasks (position
and orientation estimation), related to different visual cues, we apply a
shared convolutional backbone at two different resolutions and use two
different Transformers, one per task. The decoder's outputs are used to
classify the scene and select the respective position and orientation
embeddings from which the position and orientation vectors are regressed.
{We further extend our earlier research findings \cite{shavit2021learning}, by reframing camera pose regression as a classification-regression problem with a coarse-to-fine approach. Specifically,} by quantizing the position and
orientation domains of a scene, APR can be formulated as a classification
problem. Although classification-based
regression schemes were found to be robust \cite%
{NIPS2006_019f8b94,coral,condor2021}, they require a refinement phase to overcome quantization errors. Therefore, we propose a mixed
classification-regression architecture. {We
first classify the coarse camera location (scene) and then predict the clusters
within the scene based on the selected decoders' embeddings, which provide an initial coarse estimate of camera orientation and position. The regression network finetunes these initial estimates, by learning
the residuals with respect to the predicted quantized clusters' centers. To the best of our
knowledge, our approach is the first to suggest such a coarse-to-fine
classification-regression approach for camera pose regression.}

We evaluated our approach on two commonly reference datasets consisting of
multiple outdoor and indoor localization challenges. We show that our method
not only provides a new SOTA accuracy for \textit{multi-scene APR}
localization, but also, importantly, provides a new SOTA for \textit{%
single-scene} \textit{APRs,} outperforming our previous SOTA results \cite{shavit2021learning}. Furthermore, we show that our approach achieves
competitive results even when trained on multiple datasets with
significantly different characteristics. We further conduct multiple
ablations to evaluate the sensitivity of our model to different design
choices and analyze its scalability in terms of runtime and memory. In
summary, our main contributions are as follows:

\begin{itemize}
\item We propose a novel formulation for multi-scene absolute pose
regression using Transformers.

\item {We extend and build upon our earlier findings \cite{shavit2021learning} by approaching the APR problem through a coarse-to-fine classification-regression framework, and presenting a mixed architecture that combines both classification and regression.}

\item We experimentally demonstrate that self-attention allows aggregation
of positional and rotational image cues.

\item Our approach is shown to achieve new SOTA accuracy for both
multi-scene and single-scene APRs across contemporary outdoor and indoor
localization benchmarks.
\end{itemize}

\section{Related Work}

\label{sec:related}

\subsection{Camera Localization}

Camera pose estimation methods can be divided into several families,
depending on the inputs at inference time and on their algorithmic
characteristics.\newline
\textbf{Image Retrieval} methods learn global image descriptors for
retrieving database images that depict the vicinity of the area captured by
the query image. They are commonly employed by pose estimation methods such
as structure-based hierarchical localization pipelines \cite%
{taira2018inloc,sarlin2019coarse,dusmanu2019d2} and relative pose regression
methods \cite{balntas2018relocnet,nn-net,ding2019camnet}. IR can also be
applied to estimate the camera pose by taking the pose of the most similar
fetched image or by interpolating the poses of several visually close
images. Such approaches require both storing and searching through large
databases with pose labels. Recently, Sattler et al. \cite%
{sattler2019understanding} proposed an IR-based baseline for camera pose
regression to illustrate the limitations of APRs, as no regressor was able
to consistently surpass it on multiple localization tasks.\newline
\textbf{3D-based Localization} methods, also referred to as structure-based
methods \cite{sattler2016efficient,sattler2019understanding}, include camera
pose estimation techniques that utilize the correspondences between 2D image
positions and 3D world coordinates for camera localization with PnP and
RANSAC.\newline
\textbf{Hierarchical Structure-based pose estimation pipelines (SbP)}\cite%
{taira2018inloc,sarlin2019coarse,dusmanu2019d2} are based on a two-phase
approach, using global (IR) and local matching. Each query image to be
localized is first encoded using a CNN trained for IR, and a relatively
small set of nearest neighbors is retrieved from a large-scale image
dataset. Tentative 2D-2D correspondences are estimated by matching local
image features and then mapped into 2D-3D matches. The resulting matches are
passed to PnP-RANSAC for estimating the camera pose. Such pipelines have
been shown to achieve SOTA accuracy on large-scale benchmarks with
challenging conditions \cite{taira2018inloc,sarlin2019coarse}. A different
body of works directly regresses the 3D scene coordinates from 2D positions
in the image. Brachmann and Rother derived the DSAC \cite{DSAC} and the
follow-up DSAC++ \cite{DSAC++} schemes, where a CNN architecture is trained
to estimate the 3D locations of the pixels in the query image, in order to
establish 2D-3D correspondences for estimating the camera pose with
PnP-RANSAC. These {Scene Coordinate Regression (SCR) }methods require the query
image and the intrinsics of the query camera as input, and achieve SOTA
accuracy comparable to {SbP methods}. Similarly to
single-scene APRs, a model must be trained per scene.\newline
\textbf{Relative Pose Regression} methods typically combine camera pose
regression with an IR scheme. The absolute camera pose is computed by first
estimating the \textit{relative} motion (translation and rotation) between
the query image and a set of reference images, for which the ground truth
pose is known. An IR scheme is applied to retrieve a set of nearest-neighbor
images, and a relative motion regression is computed separately between the
query image and each of the retrieved images, followed by pose
interpolation. The learning thus focuses on regressing the relative pose
given a pair of images \cite{balntas2018relocnet,nn-net,ding2019camnet}.
These relative pose regressors (RPR) have been shown to generalize better
than APR and improve accuracy in small-scale indoor benchmarks \cite%
{ding2019camnet}. Rather than using (encoded) images with pose
labels, Aha et al. \cite{saha2018improved} suggested employing anchor points
that are uniformly distributed throughout the scene. The proposed method,
named AnchorNet, predicts which anchor points appear in the query image
along with their relative location, allowing one to compute an anchor-based
absolute pose estimate. The query pose is then computed as a weighted
average of all poses. Unlike other RPRs, AnchorNet is a single-scene
approach and is trained per scene. Some RPRs rely on sequential acquisition
of images over time \cite{valada2018deep,radwan2018vlocnet++}, while
combining relative and absolute regression, achieving an improved pose
accuracy \cite{valada2018deep,radwan2018vlocnet++}.\newline
\textbf{Absolute Pose Regression} was first proposed by \cite%
{kendall2015posenet} to directly regress the position and orientation of the
camera, given the input image, by attaching an MLP head to a GoogLeNet
backbone. The resulting architecture, named PoseNet, was much less accurate
than the 3D-based methods, but enabled pose estimation using a single forward
pass. In order to improve the localization accuracy, contemporary APRs
studied different CNN backbones \cite{naseer2017deep,shavitferensirpnet} and
MLP heads \cite{naseer2017deep}. Overfitting was addressed by averaging the
predictions of multiple models with randomly dropped activations \cite%
{kendall2016modelling}, or by reducing the dimensionality of the global
image encoding using Long-Short-Term-Memory (LSTM) layers \cite%
{walch2017image}. Other work focused on the loss formulation for absolute
pose regression to allow adaptive weighting of position- and orientation-associated errors. Kendall et al. \cite{kendall2017geometric} suggested optimizing the parameters that balance both losses to improve
accuracy and avoid manual fine-tuning. This formulation was adopted by many
pose regressors. Alternative orientation representations were also proposed
to improve pose loss \cite{brahmbhatt2018geometry}. The use of additional
sensors, such as inertial sensors, was also suggested to improve
localization accuracy \cite{brahmbhatt2018geometry}. More recently, Wang et
al. \cite{wang2020atloc} proposed using attention to guide the regression
process by applying self-attention to the output of the CNN backbone. The
new attention-based representation was used to regress the pose with an MLP
head. Although many modifications were proposed to the architecture and loss
originally formulated by Kendall et al., the main paradigm remained the
same: (1) employing a CNN backbone to output a global latent vector, which
is used for absolute pose regression, and (2) training a separate model per
scene. \newline
\textbf{Multi-Scene Absolute Pose Regression} methods aim to extend the
absolute pose regression paradigm for learning a \textit{single} model on
\textit{multiple} scenes. Blanton et al. proposed Multi-Scene PoseNet
(MSPN), a novel approach to multi-scene absolute pose regression \cite%
{blanton2020extending}, where the network first classifies the particular
scene related to the input image, and then uses it to index a set of
scene-specific weights to regress the pose. An activation map of a CNN
backbone, which is shared across scenes, is used both for scene
classification and to regress the pose. A fully connected layer with SoftMax
predicts the scene and is trained via binary cross-entropy. A set of FC
layers, one per scene, is trained for absolute camera pose regression with a
set of scene-specific parameterized losses. The notion of multi-scene camera
pose estimation was also applied to 3D-based methods, which regresses the 3D
coordinates from image pixels. However, the suggested framework still
involved training multiple models (one per scene) and then selecting the
most appropriate model using a mixture-of-experts strategy \cite%
{brachmann2019expert}.

In this work, we focus on learning a \textit{single} unified deep learning
model for absolute pose regression across multiple scenes. Our method is
thus closely related to single- and multi-scene absolute pose regression,
and we compare it to leading architectures (APRs) in this field.

\subsection{Attention and Transformers}

Attention mechanisms \cite{DBLP:journals/corr/BahdanauCB14} are layers of neural networks that aggregate information from the entire input sequence.
The aggregation is often computed by a sequence-to-sequence architecture,
where the inner-products (interactions) between the two sequences are used
to compute the aggregation weights. Attention models consist of an Encoder
and Decoder. The Encoder implements self-attention that maps the input
sequence into a higher-dimensional space that is fed into the Decoder
alongside a query sequence, outputting the result sequence. Attention allows
to numerically emphasize the contribution of the task-informative image
locations, in contrast to the visual clutter. Transformers were introduced
by Vaswani et al. \cite{AttentionIsAllYouNeed} as a novel formulation of
attention-based Encoders and Decoders for sequence encoding that does not
use RNN layers such as LSTM and GRU. Transformers consist of multiple
stacked Multi-Head Attention and Feed Forward layers. As no recurrent layers
are used, the relative position and sequential order of the sequence
elements are induced by adding positional encodings to the embedded
representation. Transformers were shown to outperform RNNs in encoding long
sequences, and were applied in multiple recent works in natural language
processing (NLP) \cite{bert,radford2019language} and computer vision \cite%
{DETR,16x16}. In this work, we propose a hybrid CNN-Transformer
architecture, inspired by recent advances in visual transformers for
multi-object detection \cite{DETR}. We employ encoders to adaptively
aggregate activation maps for position and orientation regression and use
decoders to decode aggregated representations with respect to query scenes
encoding.
\begin{figure*}%
[t]
\centering
\includegraphics[width=0.8\textwidth]{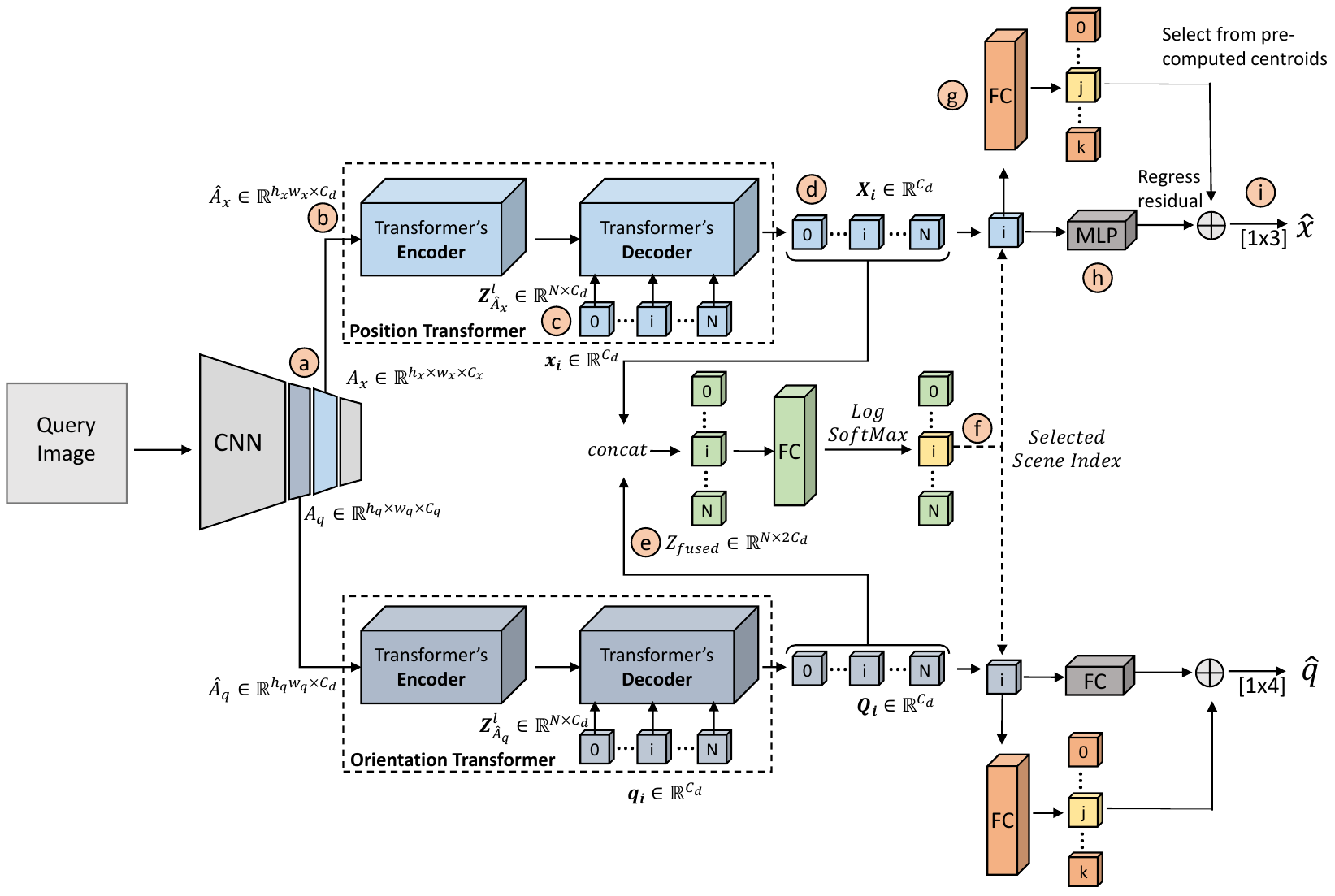}
\caption{The architecture of our proposed model.}
\label{fig:architecture}
\end{figure*}%

\section{Multi-Scene Absolute Camera Pose Regression with Transformers}
\label{sec:methods}
A single/multi-scene APR localizes the capturing camera with a forward pass
on the input image. The pose of the camera $\mathbf{p}$ is typically
represented by the tuple $<\mathbf{x},\mathbf{q}>$ where $\mathbf{x}\in
\mathbb{R}^{3}$ is the position of the camera in the world coordinates and $%
\mathbf{q}\in \mathbb{R}^{4}$ is the quaternion that encodes its 3D
orientation. Following the success of recent visual Transformers \cite%
{DETR,16x16}, we employ separate position and orientational \textit{%
Transformer Encoders} for adaptive aggregation of (flattened)
intermediate activation maps computed by a convolutional backbone. In
particular, as depicted in Figs. \ref{fig:att-enocder}-\ref{fig:att-decoder}
and Section \ref{subsubsec:image encoding}, the positional and orientational
encoders emphasize different image cues: corner- and blob-like image cues
are position-informative, in contrast to the elongated edges emphasized by
the orientation encoder.

To attend to $N$ scenes, we also apply separate positional and orientational
\textit{Transformer Decoders}, which are queried by $\left\{ \mathbf{x}%
_{i}\right\} _{1}^{N}$ and $\left\{ \mathbf{q}_{i}\right\} _{1}^{N},$ for
the position and orientation embeddings per scene, respectively. The
corresponding output sequences, $\left\{ \mathbf{X}_{i}\right\} _{1}^{N}$
and $\left\{ \mathbf{Q}_{i}\right\} _{1}^{N}$, respectively, encode the
localization parameters \emph{per scene}. This architecture is inspired by
the DETR approach \cite{DETR}, where a single activation map is queried by
multiple queries, each related to a different task. Together, the
encoder-decoder Transformer architecture allows attending
localization-informative image content while learning multiple scenes at
once. To regress the pose, the scene is first classified as described in
Section \ref{subsubsec:Scene Classfication}. By concatenating $\left\{
\mathbf{X}_{i}\right\} _{1}^{N}$ and $\left\{ \mathbf{Q}_{i}\right\}
_{1}^{N} $, the embeddings of the detected scene $\left\{ \mathbf{X}_{i},%
\mathbf{Q}_{i}\right\} $ are regressed by the MLP heads, as detailed in
Section \ref{subsubsec:Pose Classification-Regression}. The architecture of
our model is shown in Fig. \ref{fig:architecture}.

\subsection{Image Encoding using Transformers}

\label{subsubsec:image encoding}

Given an image $\mathbf{I}\in\mathbb{R}^{H\times W\times C}$, we sample a
convolutional backbone at two different resolutions and take an activation
map per regression task: $A_{\mathbf{x}}$ and $A_{\mathbf{q}}$, for the position
and orientation regression, respectively (Fig. \ref{fig:architecture}a).

In order to transform activation maps into Transformer-compatible inputs, we
follow the same sequence preparation procedure as in \cite{DETR}. An
activation map $\mathbf{A}\in \mathbb{R}^{H_{a}\times W_{a}\times C_{a}}$ is
first converted to a sequential representation $\widehat{\mathbf{A}}\in
\mathbb{R}^{{H_{a}}\cdot W_{a}\times C_{d}}$ using a $1\times 1$ convolution
(projecting to dimension $C_{d}$) followed by flattening (Figure \ref%
{fig:architecture}b). Each position in the activation map is further
assigned with a learned encoding to preserve the spatial information of each
location. In order to reduce the number of parameters, two one-dimensional
encodings are separately learned for the $X$,$Y$ axes. Specifically, for an
activation map $\mathbf{A}$ we define the sets of positional embedding
vectors $\mathbf{E}_{u}\in \mathbb{R}^{\left( W_{a}\right) \times C_{d}/2}$
and $\mathbf{E}_{v}\in \mathbb{R}^{\left( H_{a}\right) \times C_{d}/2}$,
such that a spatial position $\left( i,j\right) ,$ $i\in 1..H_{a}$, $j\in
1..W_{a}$, is encoded by concatenating the two corresponding embedding
vectors:%
\begin{equation}
\mathbf{E}_{pos}^{i,j}=%
\begin{bmatrix}
\mathbf{E}_{u}^{j} \\
\mathbf{E}_{v}^{i}%
\end{bmatrix}
\in \mathbb{R}^{C_{d}}.
\end{equation}%
The processed sequence, serving as input to the Transformer is thus given
by:
\begin{equation}
\mathbf{Z}_{\widehat{A}}^{0}=\widehat{\mathbf{A}}+\mathbf{E_{A}}\in \mathbb{R%
}^{{H_{a}}\cdot W_{a}\times C_{d}},
\end{equation}%
where $\mathbf{E_{A}}$ is the positional encoding of $A$. This processing is
applied separately for each of the two activation maps (for the position and
orientation Transformers, respectively). We use the Transformer architecture
described in \cite{DETR}, with standard Encoder and Decoders modified to add
the positional encoding at each attention layer. A Transformer Encoder is
composed of $L$ identical layers, each consisting of multi-head attention
(MHA) and multilayer perceptron (MLP) modules. Each layer $l$, $l=1..L$,
performs the following computation by applying a LayerNorm (LN) \cite%
{ba2016layer} before each module and adding back the input with residual
connections:
\begin{equation}
\mathbf{Z}_{\widehat{A}}^{l^{\prime }}=MHA(LN(\mathbf{Z}_{\widehat{A}%
}^{l-1}))+\mathbf{Z_{\widehat{A}}^{l-1}}\in \mathbb{R}^{{H_{a}}\cdot
W_{a}\times C_{d}}  \label{eq:encoder-mha}
\end{equation}%
\begin{equation}
\mathbf{Z}_{\widehat{A}}^{l}=MLP(LN(\mathbf{Z}_{\widehat{A}}^{l^{\prime }}))+%
\mathbf{Z_{\widehat{A}}^{l^{\prime }}}\in \mathbb{R}^{{H_{a}}\cdot
W_{a}\times C_{d}}  \label{eq:encoder-mlp}
\end{equation}%
At the final layer, $L$, the output is passed through an additional
normalization:
\begin{equation}
\mathbf{Z}_{\widehat{A}}^{L}=LN(\mathbf{Z}_{\widehat{A}}^{L}).
\end{equation}%
{In our model, $A_{\mathbf{x}}$ and $A_{\mathbf{q}}$ are passed to a separate Transformer Encoders which apply the aforementioned mechanism. We denote the outputs $\mathbf{Z}_{\widehat{A_{x}}}^{L}$ and $\mathbf{Z}_{\widehat{A_{q}}}^{L}$, respectively (Fig. \ref{fig:architecture}c).}

\subsection{Scene Classification}

\label{subsubsec:Scene Classfication}

Given a dataset with $N$ scenes, the Transformer Decoders first apply self-attention, as in Eq.~\ref{eq:encoder-mha}, to the two learned query sequences $\left\{ \mathbf{x}_{i}\right\} _{1}^{N}$ and $\left\{ \mathbf{q}%
_{i}\right\} _{1}^{N}$ (Fig. \ref{fig:architecture}d), for the position and
orientation decoders, respectively. Equations~\ref{eq:encoder-mha}-\ref%
{eq:encoder-mlp} are then applied again, but this time computing
encoder-decoder attention instead of self-attention. Unlike the earlier
autoregressive decoders \cite{AttentionIsAllYouNeed}, this architecture
outputs predictions in parallel for all positions. We refer the reader to
\cite{AttentionIsAllYouNeed,DETR} for detailed definitions of the MHA
operation and parallel decoding.

The Transformers Decoders output the sequences $\left\{ \mathbf{X}%
_{i}\right\} _{1}^{N}$ and $\left\{ \mathbf{Q}_{i}\right\} _{1}^{N}$,
encoding each scene with a latent embedding (Fig. \ref{fig:architecture}e).
Since a query image corresponds to a single scene from which the image was
taken, the respective latent embedding need to be selected. In order to
select the corresponding scene, we append the outputs of the two
transformers (Fig. \ref{fig:architecture}f) as $\left\{ \mathbf{Z}%
_{i\_fused}\right\} _{1}^{N},$ such that
\begin{equation}
\mathbf{Z}_{i\_fused}=%
\begin{bmatrix}
\mathbf{X}_{i} \\
\mathbf{Q}_{i}%
\end{bmatrix}%
\in \mathbb{R}^{2C_{d}},  \label{equ:class scene}
\end{equation}%
and pass them through a fully connected layer followed by Log SoftMax. The
embedding vectors $\left\{ \mathbf{X}_{i},\mathbf{Q}_{i}\right\} $
corresponding to the maximal probability of scene classification are then
chosen (Fig. \ref{fig:architecture}g).

\subsection{Pose Classification-Regression}

\label{subsubsec:Pose Classification-Regression}

The selected Transformers outputs $\left\{ \mathbf{X}_{i},\mathbf{Q}%
_{i}\right\} $ are used to refine the location of the camera, from
scene level to cluster level, and to perform residual regression. During
training, we precompute $K$ position and orientation centroids $%
\{c_{k}^{x},c_{k}^{q}\}_{1}^{K}$, respectively, for each scene using the $K$%
-means algorithm \cite{lloyd1982least}. {Given a query image, a fully
connected layer and SoftMax are applied to $\left\{ \mathbf{X}_{i},\mathbf{Q}%
_{i}\right\} $ to select the respective centroids $%
\{c_{k_{0}}^{x},c_{k_{0}}^{q}\}$ (Fig. \ref{fig:architecture}h). To further
regress the residuals of $\left\{ \mathbf{X}_{i},\mathbf{Q}_{i}\right\} $,
we apply two MLP heads with a single hidden layer and gelu non-linearity
(Fig. \ref{fig:architecture} i). The first is applied to regress the
position refinement $\mathbf{\Delta x}$, while the other MLP refines the
orientation by $\mathbf{\Delta q}$. The position and orientation vectors
(Fig. \ref{fig:architecture}j), $\mathbf{x}$ and $\mathbf{q}$, are thus
given by%
\begin{equation}
\mathbf{x}\mathbf{=}c_{k_{0}}^{x}+\mathbf{\Delta x,}\text{ }\mathbf{q}%
\mathbf{=}c_{k_{0}}^{q}+\mathbf{\Delta q.}
\end{equation}
}

\subsection{Multi-Scene Camera Pose Loss}

We train our model to minimize both the position loss $L_{\mathbf{x}}$ and
the orientation loss $L_{\mathbf{q}}$, with respect to a ground truth pose $%
\mathbf{p}_{0}=<\mathbf{x}_{0},\mathbf{q}_{0}>$, given by:
\begin{equation}
L_{\mathbf{x}}=||\mathbf{x}_{0}-\mathbf{x}||_{2}  \label{equ:position loss}
\end{equation}%
\begin{equation}
L_{\mathbf{q}}=||\mathbf{q_{0}}-\frac{\mathbf{q}}{||\mathbf{q}||}||_{2}
\label{equ:orientation loss}
\end{equation}
where $q$ is normalized to a unit norm quaternion to ensure that it is a
valid orientation encoding. We combine the two losses using the camera pose
loss formulation suggested by Kendall et al. \cite{kendall2017geometric}:
\begin{equation}
L_{\mathbf{p}}=L_{\mathbf{x}}\exp(-s_{\mathbf{x}})+s_{\mathbf{x}}+L_{\mathbf{%
q}}\exp(-s_{\mathbf{q}})+s_{\mathbf{q}}  \label{equ:learnable pose loss}
\end{equation}
where $s_{\mathbf{x}}$ and $s_{\mathbf{q}}$ are learned parameters that
control the balance between the two losses. As our model is also required to
classify the scene from which the image was taken and the centroid within
each scene, we further add the Negative Log-likelihood (NLL) loss term,
computed with respect to the ground truth scene index $s_{0}$ and the ground
truth position and orientation centroids $\mathbf{c}_{x,0}$ and $\mathbf{c}%
_{q,0}$, respectively. Given an estimated pose $p$ and the log-probability
distributions $s$, $c_{x}$ and $c_{q}$ of the predicted scene, position and
orientation centroids, our overall loss is given by:%
\begin{multline}
L_{\mathbf{multi-scene}}=L_{\mathbf{p}}+NLL(\mathbf{s},\mathbf{s_{0}})+ \\
NLL(\mathbf{c_{x}},\mathbf{c_{x,0}})+NLL(\mathbf{c_{q}},\mathbf{c_{q,0}})
\label{equ:multi-scene-loss}
\end{multline}

\subsection{Implementation Details}

Our model is implemented in PyTorch \cite{paszke2019pytorch}. We use a
pre-trained EfficientNet-B0 \cite{pmlr-v97-tan19a}, and take $A_{\mathbf{x}}$
and $A_{\mathbf{q}}$ at two different resolutions: $A_{\mathbf{x}}\in\mathbb{%
R}^{14\times14\times112}$ and $A_{\mathbf{q}}\in\mathbb{R}%
^{28\times28\times40}$. We set $C_{d}=256$ for the dimension of inputs of
the Transformer components. All encoders and decoders consist of six layers
with gelu nonlinearity and with a dropout of $p=0.1$. Each layer (encoder /
decoder) uses four MHA heads and a two-layer MLP with a hidden dimension $%
C_{h}=C_{d}$.\newline
The two MLP heads, regressing the position and orientation residual vectors,
respectively, expand the decoder dimension to $%
\mathbb{R}
^{1024}$ with a single hidden layer. Our code is publicly available\footnote{%
https://github.com/yolish/c2f-ms-transformer}, providing the model
implementation along with a training and evaluation framework, detailed
configuration files, and pre-trained models.

\section{Experimental Results}
\subsection{Experimental Setup}
\textbf{Datasets.} We evaluate our approach using the Cambridge Landmarks
\cite{kendall2015posenet} and 7Scenes \cite{glocker2013real} datasets, which
are commonly used to evaluate pose regression methods. The Cambridge
Landmarks dataset consists of six medium-sized scenes ($\sim 900-5500m^{2}$)
set in an urban environment. For our comparative analysis, we consider four
scenes that are typically benchmarked by APRs. The 7Scenes dataset includes
seven small-scale scenes ($\sim 1-10m^{2}$) set in an office indoor
environment.

\textbf{Training Details.} We optimize our model to minimize the loss in Eq. %
\ref{equ:multi-scene-loss} using Adam, with $\beta _{1}=0.9$, $\beta
_{2}=0.999$ and $\epsilon =10^{-10}$. The loss parameters (Eq. \ref%
{equ:learnable pose loss}) are initialized as in \cite{valada2018deep}.
Throughout all experiments, we use a batch size of $8$ and an initial
learning rate of $\lambda =10^{-4}$. For each dataset, we initialize our
model from the respective pre-trained MS-Transformer model \cite%
{shavit2021learning}, which we extend in this work. We use $k=4$ for the
number of position and orientation clusters for the CambridgeLandmarks
dataset and $k=1$, $k=2$ for the number of position and orientation clusters
for the 7Scenes dataset. At train time, the decoder output is selected using
the ground truth scene index, and the estimated scene index is used only for
evaluating the NLL loss. During inference, the scene index is unknown, and
we rely on the prediction of our model, taking the index with the highest
(log) probability. The same logic is applied for training the classifiers of
the position and orientation centroids. Note that instead of training a
model per scene (typically with scene-specific training hyperparameters) as
in single APRs, here we train a single model for \emph{all} the scenes
together. All experiments reported in this paper were performed on an 8GB
NVIDIA GeForce GTX 1080 GPU. We follow the augmentation procedure described
in \cite{kendall2015posenet}. During training, the images are first rescaled
so that the smaller edge is resized to $256$ pixels and then randomly
cropped to a $224\times 224$-sized image. Additionally, brightness,
contrast, and saturation are randomly jittered. At test time, the center
crop is taken after rescaling without any further augmentations. For the
7Scenes dataset, we train with the aforementioned augmentation scheme for $%
30 $ epochs and reduce the learning rate by half every $10$ epochs.{%
For the Cambridge Landmarks dataset, we train for $550$ epochs and reduce
the learning rate by half every $200$ epochs. For both datasets, we
fine-tune the respective models from \cite{shavit2021learning}. Table \ref%
{tb:training_time} shows the training times for these models.}
\begin{table}[tbh]
\centering
\caption{The Training Time of our model.}
\label{tb:training_time}\centering
\begin{tabular}{lccc}
\toprule
\textbf{Dataset} & \textbf{\#Images} & \textbf{\#Epochs} & \textbf{Training
Time } \\
&  &  & \textbf{[hours]} \\ \midrule
{Cambridge} & 3,834 & 550 & 24 \\
{7Scenes} & 26,000 & 30 & 10 \\ \bottomrule
\end{tabular}

\end{table}

\subsection{Comparative Localization Analysis}

\label{sec:comparative_analysis}
Our proposed coarse-to-fine approach (c2f-MS-Transformer) is {a \textit{multiscene} absolute pose regression (MS-APR) paradigm}.
As such, we compare it to single-scene {(SS-APR) and MS-APRs}  using the
CambridgeLandmarks and 7Scenes datasets (Table \ref{tb:cambridge_rank} and Table \ref{tb:7scenes_rank}, respectively). {For our approach, we report the results when training from scratch and when initializing from the respective MS-Transformer \cite{shavit2021learning} model.}
{To provide a wider experimental overview and context, we also include the results of representative methods from other classes of localization schemes (see Section \ref{sec:related}), namely: SbP, SCR, Sequence-based (Seq.) and IR. We report the median position and orientation errors and the average of errors across scenes.}  {The DSAC* approach (SCR family) achieves the overall SOTA accuracy by a significant margin} for the CambridgeLandmarks and 7Scenes datasets, where the closest second is the sequential RPR-based VLocNet++ \cite{radwan2018vlocnet++}.
{Within the APR family, our method is the most accurate across datasets}.
MSPN \cite{blanton2020extending}, is, to the best
of our knowledge, the only other MS-APR approach to date. Since it was trained on a different combination of scenes from the CambridgeLandmarks dataset, we
applied the best performing model reported by the authors on this dataset
\cite{blanton2020extending}. Our method consistently outperforms MSPN across
outdoor and indoor scenes, reducing both position and orientation errors.
{Moreover,} compared to single- and multiple-scene APRs, the c2f-MS-Transformer achieves SOTA positional accuracy in 10 of the 11 scenes (through the CambridgeLandmarks and 7Scenes datasets).
Furthermore, our method achieves the lowest average positional error for
both indoor and outdoor localization and is the only APR approach to report
an average error below one meter in outdoor scenes. In most of the scenes
reported, c2f-MS-Transformer outperforms our previous method \cite%
{shavit2021learning} in both position and orientation accuracy.
Interestingly, the two best-performing APRs on the 7Scenes dataset (AttLoc
and our method) both use the attention mechanism for pose regression.

{We note that different localization approaches offer different trade-offs between accuracy and other attributes such as generalization and storage. Table \ref{tb:tradeoff} summarizes the key characteristics of the classes of localization schemes compared in Tables \ref{tb:cambridge_rank} and \ref{tb:7scenes_rank}. We consider whether the localization scheme can localize using only the query image (Query only) as opposed to requiring a database or a sequential acquisition (as in SbP, Seq, IR and RPR methods), whether it can localize without a 3D model (No 3D), whether it can learn multiple scenes at once (Multi-scene), generalize to unseen scenes (Generalization) and whether it provides high accuracy (sub-degree error and $<0.3$ meters and $<0.05$ meters for mid- and small-scale scenes, respectively). Multi-Scene APR approaches can operate without 3D information using the query image alone, while learning multiple scenes at once. However, they are less accurate than SbP and Sequential and SCR methods and cannot generalize to unseen scenes, as opposed to SbP, IR and RPR approaches. Tables \ref{table:model_size} and \ref{table:storage} further compare runtime, storage, and the model size of our method and other representative approaches}.

\begin{table*}%
\caption{Localization results for \textbf{Cambridge Landmarks} dataset.
We report the average median position/orientation errors in
meters/degrees and the respective rankings. The most accurate global results
are highlighted in \protect\underline{\textbf{underlined bold}}, while the most accurate APR results are highlighted in \textbf{bold}.}
\label{tb:cambridge_rank}\centering{\
\begin{tabular}{llccccc}
\toprule
& \textbf{Method} & \textbf{College} & \textbf{Hospital} & \textbf{Shops} & \textbf{St.
Mary's} & \textbf{Avg.} \\ \midrule
\multirow{2}{*}{{\rotatebox[origin=c]{90}{SbP}}} & ActiveSearch &
0.42,0.6\degree & 0.44,1.0%
\degree & 0.12,0.40\degree & 0.19,0.5\degree & 0.29,0.63\degree \\
& InLoc\cite{taira2018inloc}&0.18,0.6\degree&1.2,0.6\degree&0.48,1.0\degree&0.46,0.8\degree& 0.11,0.5\degree \\ \midrule
\multirow{3}{*}{{\rotatebox[origin=c]{90}{SCR}}}
& {DSAC \cite{DSAC}} & 0.30,0.5\degree & 0.33,0.6\degree & 0.09,0.4\degree &
0.55,1.6\degree & 0.31,0.78\degree \\
& {DSAC++ \cite{DSAC++}} & \underline{\textbf{0.18,0.3\degree}} & \underline{%
\textbf{0.20,0.3\degree}} & 0.06,\underline{\textbf{0.3\degree}} & \underline{\textbf{%
0.13,0.4\degree}} & \underline{\textbf{0.14,0.33}\degree} \\
& DSAC\textsuperscript{*}~%
\cite{9394752}~\cite{9394752} &
\underline{\textbf{0.18,0.3\degree}} & 0.21,0.4\degree & \underline{\textbf{%
0.05,0.3\degree}} & 0.15,0.5\degree & 0.15,0.4\degree \\\midrule
\multirow{3}{*}{{\rotatebox[origin=c]{90}{Seq.}}} & MapNet~\cite%
{brahmbhatt2018geometry} & 1.08,1.9\degree & 1.94,3.9\degree & 1.49,4.2%
\degree & 2.00,4.5\degree & 1.63,3.6\degree \\
& GL-Net~\cite{glnet} & 0.59,0.7\degree & 1.88,2.8\degree & 0.50,2.9\degree
& 1.90,3.3\degree & 1.22,2.4\degree \\
& {VLocNet \cite{valada2018deep}} & 0.83/1.42\degree & 1.07/2.41\degree &
0.59/3.53\degree & 0.63/3.91\degree & 0.78,2.81\degree \\ \midrule
\multirow{2}{*}{{\rotatebox[origin=c]{90}{IR}}} & VLAD~\cite{denseVLAD} &
2.80,5.7\degree & 4.01,7.1\degree & 1.11,7.6\degree & 2.31,8.0\degree &
2.56,7.1\degree \\
& VLAD+Inter~\cite{sattler2019understanding} & 1.48,4.5\degree & 2.68,4.6%
\degree & 0.90,4.3\degree & 1.62,6.1\degree & 1.67,4.9\degree \\ \midrule
\multirow{3}{*}{{\rotatebox[origin=c]{90}{RPR}}} & EssNet~\cite{essnet} &
0.76,1.9\degree & 1.39,2.8\degree & 0.84,4.3\degree & 1.32,4.7\degree &
1.08,3.4\degree \\
& NC-EssNet~\cite{essnet} & 0.61,1.6\degree & 0.95,2.7\degree & 0.70,3.4%
\degree & 1.12,3.6\degree & 0.85,2.8\degree \\
& RelocGNN\cite{9394752} & 0.48,1.0\degree & 1.14,2.5\degree & 0.48,2.5%
\degree & 1.52,3.2\degree & 0.91,2.3\degree \\ \midrule
\multirow{7}{*}{{\rotatebox[origin=c]{90}{SS-APR}}}
& PoseNet \cite%
{kendall2015posenet} & 1.92,5.40\degree & 2.31,5.38\degree & 1.46,8.08\degree
& 2.65,8.48\degree & {2.08,6.83\degree} \\
& BayesianPN \cite{kendall2016modelling} & 1.74,4.06\degree & 2.57,5.14%
\degree & 1.25,7.54\degree & 2.11,8.38\degree & {1.91,6.28\degree} \\
& LSTM-PN \cite{walch2017image} & 0.99,3.65\degree & 1.51,4.29\degree &
1.18,7.44\degree & 1.52,6.68\degree & {1.30,5.57\degree} \\
& SVS-Pose \cite{naseer2017deep} & 1.06,2.81\degree & 1.50,4.03\degree &
0.63,5.73\degree & 2.11,8.11\degree & {1.32,5.17\degree} \\
& GPoseNet \cite{cai2019hybrid} & 1.61,2.29\degree & 2.62,3.89\degree &
1.14,5.73\degree & 2.93,6.46\degree & {2.07,4.59\degree} \\
& PoseNetLearn \cite{kendall2017geometric} & 0.99,1.06\degree & 2.17,2.94%
\degree & 1.05,3.97\degree & 1.49,3.43\degree & {1.42,2.85\degree} \\
& GeoPoseNet \cite{kendall2017geometric} & 0.88,\textbf{1.04}\degree &
3.20,3.29\degree & 0.88,3.78\degree & 1.57,\textbf{3.32}\degree & {1.63,2.86%
\degree} \\
& MapNet \cite{brahmbhatt2018geometry} & 1.07,1.89\degree & 1.94,3.91\degree
& 1.49,4.22\degree & 2.00,4.53\degree & {1.62,3.64\degree} \\
& IRPNet \cite{shavitferensirpnet} & 1.18,2.19\degree & 1.87,3.38\degree &
0.72,3.47\degree & 1.87,4.94\degree & {1.41,3.50\degree} \\ \midrule
\multirow{4}{*}{{\rotatebox[origin=c]{90}{MS-APR}}}
& MSPN \cite{blanton2020extending} & 1.73/3.65\degree & 2.55/4.05\degree & 2.92/7.49\degree &
2.67/6.18\degree &  2.47/5.34\degree \\
& MS-Trans\cite{shavit2021learning} & 0.83,1.47\degree & 1.81,\textbf{2.39}%
\degree & 0.86,3.07\degree & 1.62,3.99\degree & 1.28,\textbf{2.73}\degree \\
& \textbf{c2f-MS-Trans w/o init (ours)} & 0.82,2.34\degree &	2.10,2.79\degree&0.90,3.21\degree&1.27/3.69\degree&1.27,\textbf{3.01}\\
& \textbf{c2f-MS-Trans (ours)} & \textbf{0.70},2.69\degree & \textbf{1.48},2.94%
\degree & \textbf{0.59,2.88}\degree & \textbf{1.14},3.88\degree & {\textbf{%
0.98}/3.10\degree} \\ \bottomrule
\end{tabular}
}
\end{table*}%

\begin{table*}[tbh]
\caption{Localization results for \textbf{7Scenes} dataset. We report
the average of median position/orientation errors in meters/degrees and the
respective rankings. The most accurate global results are highlighted in
\protect\underline{\textbf{underlined bold}}, while the most accurate APR
results are highlighted in \textbf{bold}. }
\label{tb:7scenes_rank}%
\centering
\begin{tabular}{clcccccccc}
\toprule
&  & \textbf{Chess} & \textbf{Fire} & \textbf{Heads} & \textbf{Office} &
\textbf{Pumpkin} & \textbf{Kitchen} & \textbf{Stairs} & \textbf{Avg.} \\
\midrule
\multirow{2}{*}{{\rotatebox[origin=c]{90}{SbP}}} & {Active Search \cite{sattler2016efficient}} & 0.04,2.0\degree & 0.03,1.5%
\degree & 0.02,1.5\degree & 0.09,3.6\degree & 0.08,3.1\degree & 0.07,3.4%
\degree & 0.03,2.2\degree & 0.05,2.47\degree \\
& InLoc\cite{taira2018inloc}& 0.03,1.05\degree & 0.03,1.07\degree & 0.02,1.16\degree &0.03,1.05\degree&0.05,1.55\degree  & 0.04,1.31\degree&0.09,2.47\degree & 0.04,1.44\degree\\ \midrule
\multirow{3}{*}{{\rotatebox[origin=c]{90}{SCR}}}
& {DSAC \cite{DSAC}} & \underline{\textbf{0.02}},0.7\degree &0.03,1.0\degree&0.02,1.30\degree&0.03,	1.0\degree&0.05,1.30\degree&0.05,1.5\degree&1.90,49.4\degree&0.30,8.03\degree
\\
& {DSAC++\cite{DSAC++}} &
\underline{\textbf{0.02}},\underline{\textbf{0.5}}\degree & {0.02}
\underline{,\textbf{0.9\degree}} & \underline{\textbf{0.01,0.8\degree}} &
0.03,\underline{\textbf{0.7\degree}} & 0.04,\underline{\textbf{1.1\degree}}
& 0.04,\underline{\textbf{1.1\degree}} & 0.09,2.6\degree
& 0.04,\underline{\textbf{1.10\degree}} \\
 & DSAC\textsuperscript{*}~%
\cite{9394752} & \underline{\textbf{0.02}},1.1\degree & {0.02},1.0\degree &
\underline{\textbf{0.01}},1.8\degree & 0.03,1.2\degree & 0.04,1.4\degree &
\underline{\textbf{0.03}},1.7\degree & 0.04,1.4\degree & 0.03,1.37\degree \\\midrule
\multirow{5}{*}{{\rotatebox[origin=c]{90}{Seq.}}} & LsG~\cite{lsg} & 0.09,3.3%
\degree & 0.26,10.9\degree & 0.17,12.7\degree & 0.18,5.5\degree & 0.20,3.7%
\degree & 0.23,4.9\degree & 0.23,11.3\degree & 0.19,7.47\degree \\
& MapNet\cite{brahmbhatt2018geometry} & 0.08,3.3\degree & 0.27,11.7\degree &
0.18,13.3\degree & 0.17,5.2\degree & 0.22,4.0\degree & 0.23,4.9\degree &
0.30,12.1\degree & 0.21,7.78\degree \\
& GL-Net\cite{glnet} & 0.08,2.8\degree & 0.26,8.9\degree & 0.17,11.4\degree
& 0.18,13.3\degree & 0.15,2.8\degree & 0.25,4.5\degree & 0.23,8.8\degree &
0.19,7.50\degree \\
& {VLocNet \cite{valada2018deep}} & 0.04,1.71\degree & 0.04,5.34\degree &
0.05,6.64\degree & 0.04,1.95\degree & 0.04,2.28\degree & 0.04,2.20\degree &
0.10,6.48\degree & 0.05,3.80\degree \\
& {VLocNet++ \cite{radwan2018vlocnet++}} & \underline{\textbf{0.02}},1.44%
\degree & \underline{\textbf{0.01}},1.39\degree & 0.02,0.99\degree &
\underline{\textbf{0.02}},1.14\degree & \underline{\textbf{0.02}},1.45\degree
& \underline{\textbf{0.03}},2.27\degree & \underline{\textbf{0.02}},\underline {\textbf{1.08%
\degree}} & \underline{\textbf{0.02}},1.39\degree \\ \midrule
\multirow{2}{*}{{\rotatebox[origin=c]{90}{IR}}} & VLAD~\cite{denseVLAD} &
0.21,12.5\degree & 0.33,13.8\degree & 0.15,14.9\degree & 0.28,11.2\degree &
0.31,11.2\degree & 0.30,11.3\degree & 0.25,12.3\degree & 0.26,12.46\degree
\\
& VLAD+Inter\cite{sattler2019understanding} & 0.18,10.0\degree & 0.33,12.4%
\degree & 0.14,14.3\degree & 0.25,10.1\degree & 0.26,9.4\degree & 0.27,11.1%
\degree & 0.24,14.7\degree & 0.24,11.71\degree \\ \midrule
\multirow{6}{*}{{\rotatebox[origin=c]{90}{RPR}}} & NN-Net~\cite{nn-net} &
0.13,6.5\degree & 0.26,12.7\degree & 0.14,12.3\degree & 0.21,7.4\degree &
0.24,6.4\degree & 0.24,8.0\degree & 0.27,11.8\degree & 0.21,9.30\degree \\
& RelocNet\cite{balntas2018relocnet} & 0.12,4.1\degree & 0.26,10.4\degree &
0.14,10.5\degree & 0.18,5.3\degree & 0.26,4.2\degree & 0.23,5.1\degree &
0.28,7.5\degree & 0.21,6.73\degree \\
& EssNet~\cite{essnet} & 0.13,5.1\degree & 0.27,10.1\degree & 0.15,9.9\degree
& 0.21,6.9\degree & 0.22,6.1\degree & 0.23,6.9\degree & 0.32,11.2\degree &
0.22,8.03\degree \\
& NC-EssNet~\cite{essnet} & 0.12,5.6\degree & 0.26,9.6\degree & 0.14,10.7%
\degree & 0.20,6.7\degree & 0.22,5.7\degree & 0.22,6.3\degree & 0.31,7.9%
\degree & 0.21,7.50\degree \\
& RelocGNN\cite{9394752} & 0.08,2.7\degree & 0.21,7.5\degree & 0.13,8.70%
\degree & 0.15,4.1\degree & 0.15,3.5\degree & 0.19,3.7\degree & 0.22,6.5%
\degree & 0.16,5.24\degree \\
& {CamNet \cite{ding2019camnet}} & 0.04,1.73\degree & 0.03, 1.74\degree &
0.05,1.98\degree & 0.04,1.62\degree & 0.04,1.64\degree & 0.04,1.63\degree &
0.04,1.51\degree & 0.04,1.69\degree \\ \midrule
\multirow{6}{*}{{\rotatebox[origin=c]{90}{SS-APR}}} & PoseNet \cite%
{kendall2015posenet} & 0.32,8.12\degree & 0.47,14.4\degree & 0.29,12.0\degree
& 0.48,7.68\degree & 0.47,8.42\degree & 0.59,8.64\degree & 0.47,13.8\degree
& 0.44,10.44\degree \\
& BayesianPN \cite{kendall2016modelling} & 0.37,7.24\degree & 0.43,13.7%
\degree & 0.31,12.0\degree & 0.48,8.04\degree & 0.61,7.08\degree & 0.58,7.54%
\degree & 0.48,13.1\degree & 0.47,9.81\degree \\
& LSTM-PN \cite{walch2017image} & 0.24,5.77\degree & 0.34,11.9\degree &
0.21,13.7\degree & 0.30,8.08\degree & 0.33,7.00\degree & 0.37,8.83\degree &
0.40,13.7\degree & 0.31,9.85\degree \\
& GPoseNet \cite{cai2019hybrid} & 0.20,7.11\degree & 0.38,12.3\degree &
0.21,13.8\degree & 0.28,8.83\degree & 0.37,6.94\degree & 0.35,8.15\degree &
0.37,12.5\degree & 0.31,9.95\degree \\
& PoseNetLearn\cite{kendall2017geometric} & 0.14,4.50\degree & 0.27,11.8%
\degree & 0.18,12.1\degree & 0.20,5.77\degree & 0.25,4.82\degree & 0.24,5.52%
\degree & 0.37,10.6\degree & 0.24,7.87\degree \\
& GeoPoseNet\cite{kendall2017geometric} & 0.13,4.48\degree & 0.27,11.3\degree
& 0.17,13.0\degree & 0.19,5.55\degree & 0.26,4.75\degree & 0.23,\textbf{5.35}%
\degree & 0.35,12.4\degree & 0.23,8.12\degree \\
& IRPNet\cite{shavitferensirpnet} & 0.13,5.64\degree & 0.25,9.67\degree &
0.15,13.1\degree & 0.24,6.33\degree & 0.22,5.78\degree & 0.30,7.29\degree &
0.34,11.6\degree & 0.23,8.49\degree \\
& AttLoc\cite{wang2020atloc} & {0.10},\textbf{4.07}\degree & 0.25,11.4\degree
& 0.16,\textbf{11.8}\degree & 0.17,\textbf{5.34}\degree & 0.21,\textbf{4.37}%
\degree & 0.23,5.42\degree & 0.26,10.5\degree & 0.20,7.56\degree \\ \midrule
\multirow{4}{*}{{\rotatebox[origin=c]{90}{MS-APR}}}
& MSPN\cite{blanton2020extending} & \textbf{0.09}/4.76\degree & 0.29/10.5\degree &
0.16/13.1\degree & \textbf{0.16}/6.80\degree & 0.19/5.50\degree & {0.21}/6.61\degree & 0.31/11.6\degree &  0.20/7.56\degree
\\
& MS-Trans\cite{shavit2021learning} & 0.11,4.66\degree & \textbf{0.24,9.60}%
\degree & 0.14,12.2\degree & 0.17,5.66\degree & 0.18,4.44\degree & 0.17,5.94%
\degree & 0.26,8.45\degree & 0.18,\textbf{7.28}\degree \\
& \textbf{c2f-MS-Trans w/o init. (ours) } & 0.10,4.97\degree & 0.23,9.46\degree & 0.13,12.4\degree &	0.17,5.71\degree	& 0.17,4.48& 0.17,6.23\degree&0.25,	9.62\degree &	0.18,7.55\degree\\
& \textbf{c2f-MS-Trans (ours) } & {0.10},4.60\degree & \textbf{0.24},9.88%
\degree & \textbf{0.12},12.3\degree & \textbf{0.16},5.64\degree & \textbf{%
0.16},4.42\degree & \textbf{0.16},6.39\degree & \textbf{0.25,7.76}\degree &
\textbf{0.17/7.28}\degree \\ \bottomrule
\end{tabular}
\end{table*}

\begin{table*}[tbh]
\caption{The attributes of the main classes of localization schemes. We indicate that an approach has an attribute by a \checkmark sign. The '*' sign indicates that the attribute is manifested by some schemes within the class.}
\label{tb:tradeoff}%
\centering
\begin{tabular}
{cccccc}
\toprule
\textbf{Family} & \textbf{Query-only} & \textbf{No 3D} & \textbf{Multi-Scene} & \textbf{Generalization} &  \textbf{High Accuracy} \\ \midrule
SbP & & & \checkmark & \checkmark & \checkmark \\
SCR & \checkmark & * & & & \checkmark \\
Seq &  & \checkmark & \checkmark & \checkmark & \checkmark \\
IR & & \checkmark & \checkmark & \checkmark & \\
RPR &  & \checkmark & \checkmark & \checkmark & \\
Single-Scene APR  & \checkmark & \checkmark & &  \\
Multi-Scene APR  & \checkmark & \checkmark & \checkmark &  \\
\bottomrule
\end{tabular}
\end{table*}

We can further extend the notion of multi-scene learning to multi-dataset
learning, where a single model is trained on completely separate datasets,
potentially having different challenges and attributes. To evaluate the
effect of such an extension, we jointly trained our model on both the
7Scenes and Cambridge Landmarks datasets. Table \ref{tb:multi_dataset} shows
the average pose error per dataset for a \textit{state-of-the-art}
single-scene APR \cite{kendall2017geometric} and our model, trained in
multi-scene and in multi-dataset modes. Although some degradation is
observed when both datasets are jointly trained, our model still maintains
competitive performance and outperforms the single-scene model. This is
despite the fact that the two datasets represent significantly different
environments and challenges (medium-scale outdoor versus small-scale
indoor). We also evaluated the ability of our model to correctly classify
the scene of the input query image. Our model achieves an average accuracy
of $98.9\%$ (across scenes) allowing for a reliable selection of the decoder
output, which is essential for regressing the pose.
\begin{table}[tbh]
\caption{Localization results with single-scene, multi-scene and
multi-dataset learning. We report the average of median position/orientation
errors in meters/degrees for the CambridgeLandmarks and 7Scenes datasets.}
\label{tb:multi_dataset}\centering%
\begin{tabular}{lcc}
\toprule
\textbf{APR Method} & \textbf{CambridgeLand.} & \textbf{7Scenes} \\
& \textbf{[m/deg]} & \textbf{[m/deg]} \\ \midrule
{Single-scene \cite{kendall2017geometric}} & 1.43/2.85 & 0.24/7.87 \\ \midrule
{Multi-scene (Ours) \cite{shavit2021learning}} & 1.28/2.73 & 0.18/7.28 \\
{Multi-dataset (Ours) \cite{shavit2021learning}} & {1.50/\textbf{2.57}} & {%
0.22/6.78} \\ \midrule
{{c2f}-Multi-scene (Ours)} & {\textbf{0.98}/3.10} & {\textbf{0.17}}/7.28 \\
{{c2f}-Multi-dataset (Ours)} & {1.59/2.64} & {0.28/\textbf{6.44}} \\ \bottomrule
\end{tabular}%
\end{table}

\subsection{Attention Maps Visualization and Interpretation}

\label{sec:att-vis}

\begin{figure}[tbh]
\begin{center}
\subfigure[Input (3
scenes)]{\includegraphics[scale=0.25]{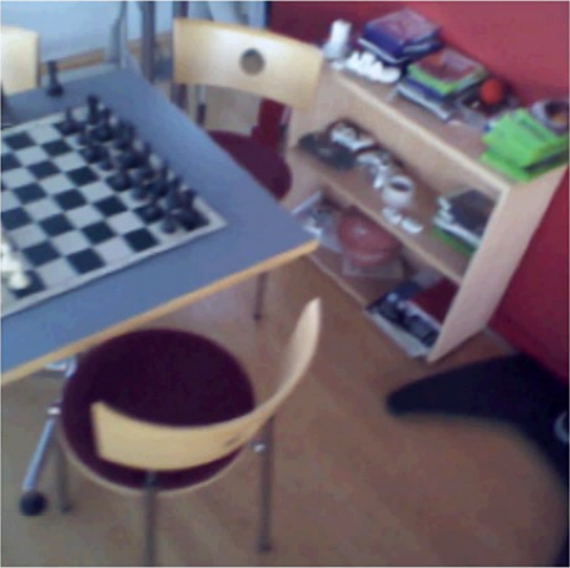}} %
\subfigure[Position]{%
\includegraphics[scale=0.25]{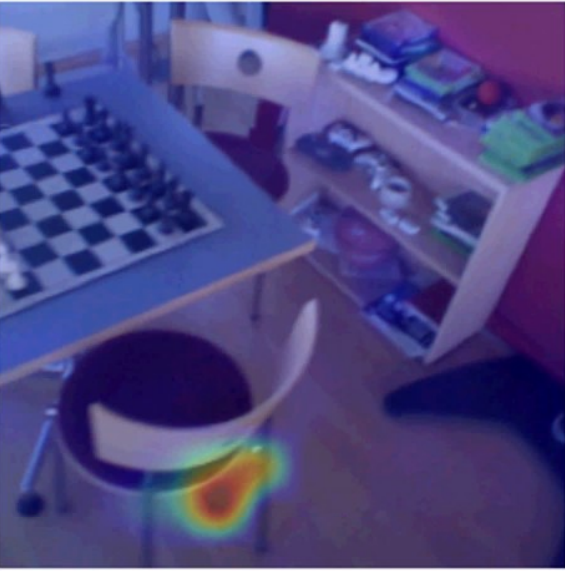}} %
\subfigure[Orientation]{%
\includegraphics[scale=0.25]{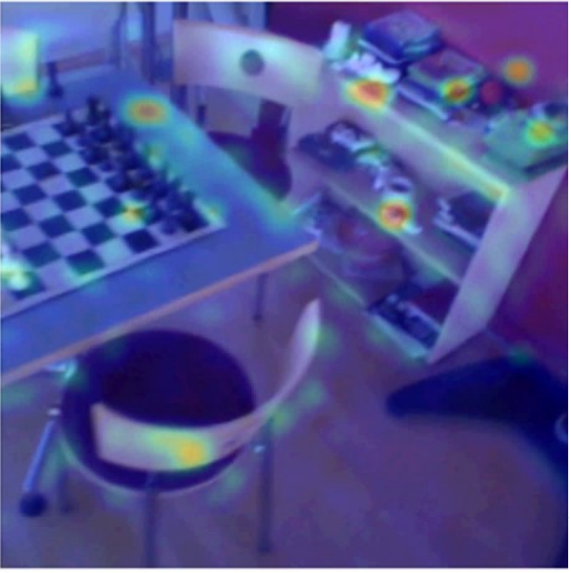}}
\subfigure[Input (7
scenes)]{\includegraphics[scale=0.25]{chess_encoder_act_maps_comp1.pdf}} %
\subfigure[Position]{%
\includegraphics[scale=0.25]{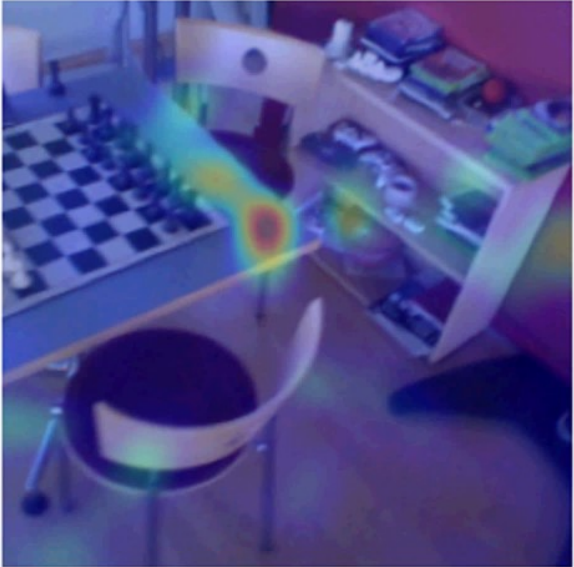}} %
\subfigure[Orientation]{%
\includegraphics[scale=0.25]{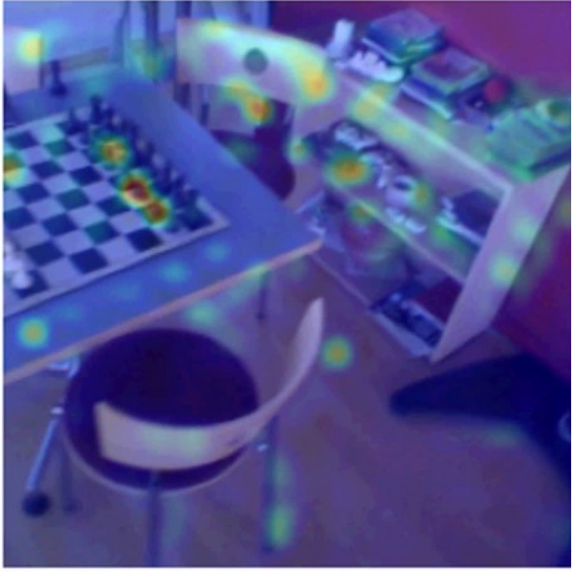}}
\end{center}
\caption{Transformer Encoder attention visualizations for a varying number
of training scenes (three and seven). As we train our scheme using more
scenes (second row), the positional attention is able to better localize
corner-like image cues (e compared to b). The orientational attention is
able to better localize elongated edges (f compared to c). }
\label{fig:att-enocder}
\end{figure}

\begin{figure}[tbh]
\begin{center}
\subfigure[Kings
C.]{\includegraphics[scale=0.18]{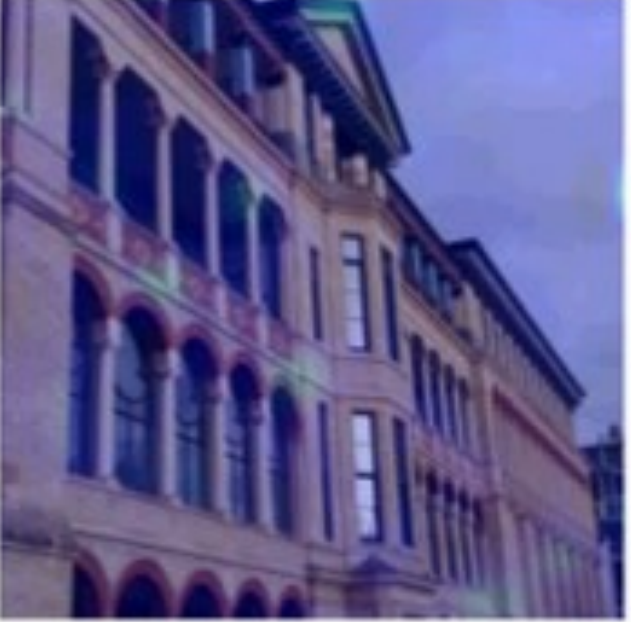}}
\subfigure[Old
Hospital]{\includegraphics[scale=0.18]{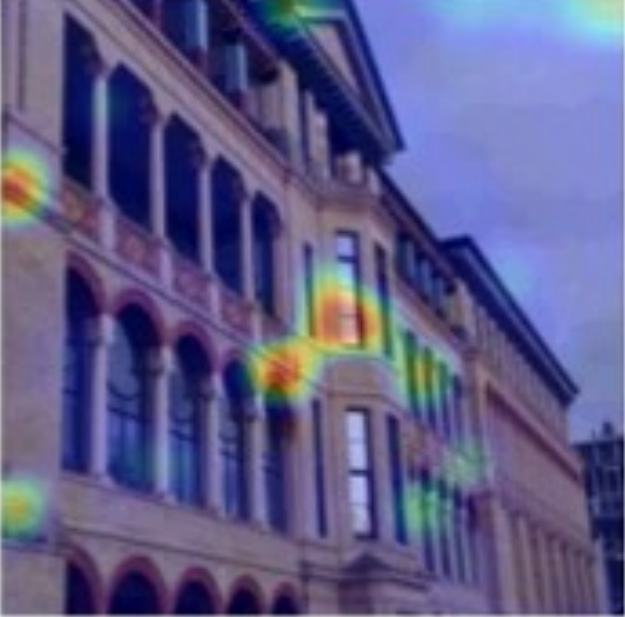}}
\subfigure[Shop
Facade]{\includegraphics[scale=0.18]{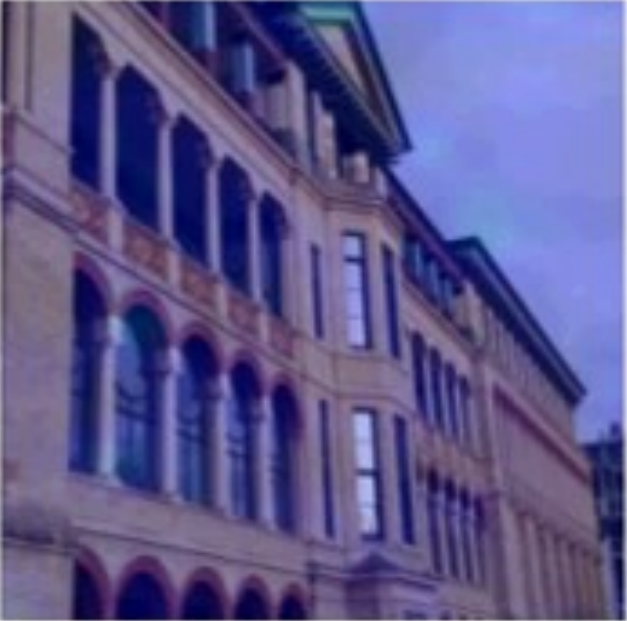}}
\subfigure[St.
Mary's]{\includegraphics[scale=0.18]{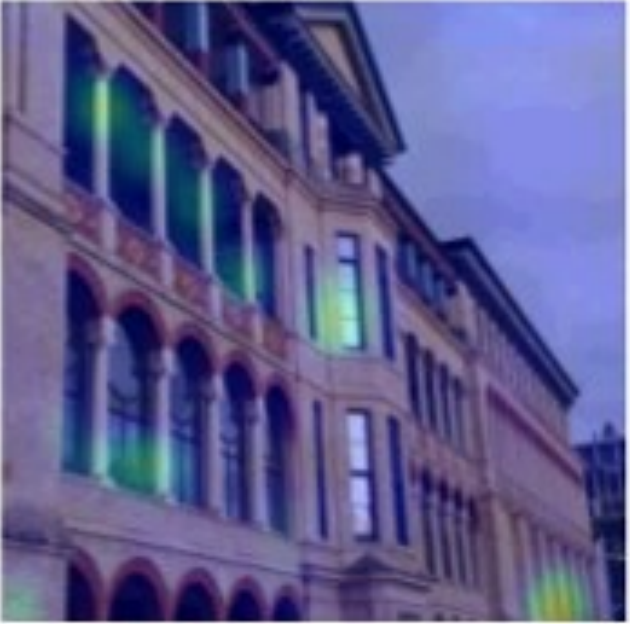}}
\end{center}
\caption{Translational Decoder attention visualization $\left\{ \mathbf{X}%
_{i}\right\} _{1}^{N}$. Each activation relates to a different scene. The
activations are due to an input image from the old Hospital scene. The
activations of the corresponding scene are notably stronger.}
\label{fig:att-decoder}
\end{figure}

In attention-based schemes, attention maps provide intuitive interpretation
and understanding of the visual cues captured by Transformer Encoders. Using
heatmaps overlaid on the input images, we visualize the upsampled attention
weights of the last encoder layer. Figure \ref{fig:att-enocder} shows the
attention map of an image taken from the \textit{Chess} scene in the 7Scenes
dataset. We show the activations when training on three and seven scenes
(top and bottom rows, respectively). Training on more (seven) scenes allows
the network to better capture informative image cues for both positional and
orientational embedding. In particular, positional attention focuses on
corners and blob-like objects, while orientational attention emphasizes
elongated edges. We also visualize the attentions $\left\{ \mathbf{X}%
_{i}\right\} _{1}^{N}$ on the output of the positional decoder for an image
from the \textit{OldHospital} scene (Fig. \ref{fig:att-decoder}). Each
activation corresponds to a particular scene. In fact, the activations
corresponding to the OldHospital scene (Fig. \ref{fig:att-decoder}b) are
significantly stronger.

\subsubsection{Visualization of Decoder Attention}

\begin{figure}[tbh]
\begin{center}
\subfigure[Kings
C.]{\includegraphics[scale=0.4]{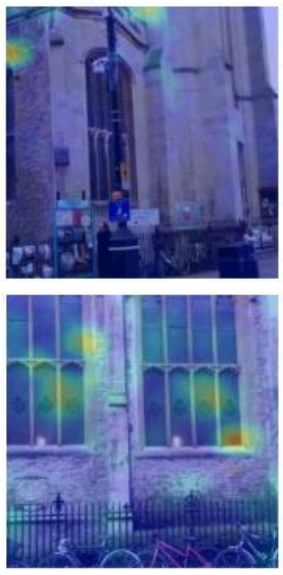}}
\subfigure[Old
Hospital]{\includegraphics[scale=0.4]{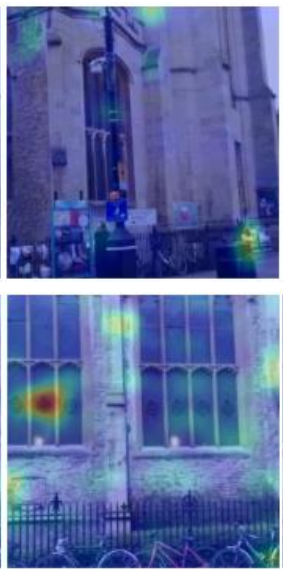}}
\subfigure[Shop
Facade]{\includegraphics[scale=0.4]{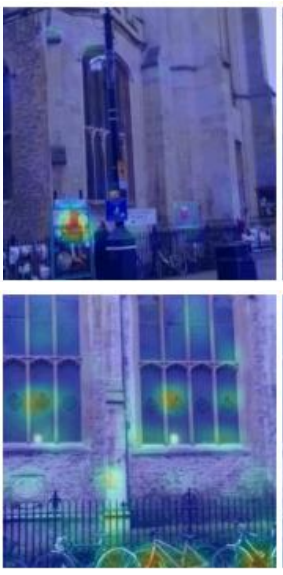}}
\subfigure[St.
Mary's]{\includegraphics[scale=0.4]{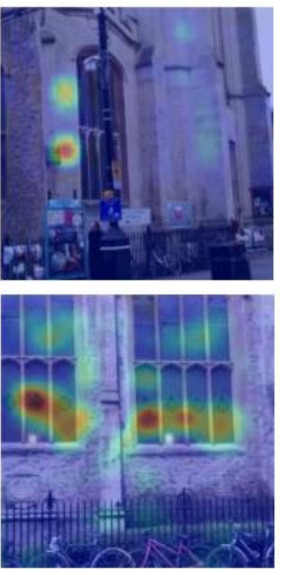}}
\end{center}
\caption{Translational Decoder attention visualization $\left\{ \mathbf{X}%
_{i}\right\} _{1}^{N}$. Each activation relates to a different scene. The
activations are due to two input images from the St Mary's scene. The
activations of the corresponding scene are notably stronger.}
\label{fig:att-decoder-2}
\end{figure}

In order to gain additional insights into the scene-specific features
learned by our model, we further visualize the attentions $\left\{ \mathbf{X}%
_{i}\right\} _{1}^{N}$ on the output of the positional decoder in two images
from the \textit{St. Mary's} scene. In addition, we measure and rank the
decoder outputs by summing over the corresponding attention map. Indeed, the
strongest response is obtained in the output corresponding to that scene
(Fig. \ref{fig:att-decoder-2}d). Interestingly, the activations related to
the \textit{ShopFacade} (Fig. \ref{fig:att-decoder-2} c) scene are focused
on the lower part of the input images, which typically includes the key
features in images from this scene.

\subsection{Ablation Study}
To study the effect of different architectural design choices, we conducted
multiple ablation experiments on the 7Scenes dataset (Tables \ref%
{table:ablation-cnn}-\ref{table:ablation-transformer-dims}). In each
experiment, we start with the architecture used for our comparative analysis, {without initialization},
(Section \ref{sec:comparative_analysis}) and modify a single algorithmic
component / hyperparameter. {Our ablation study focuses on three main
aspects of our approach: (a) the derivation of activation maps (backbone and
resolution) (b) the transformer architecture and (c) coarse-to-fine method
(number of clusters). For the ablation of the number of clusters, w}e used
the same backbone and Transformers configuration as in \cite%
{shavit2021learning}, to allow accurate comparisons with the proposed
scheme. We computed the median position and orientation errors for each
scene and reported the average between scenes. \newline
\textbf{Convolutional Backbone.} We consider three convolutional encoders
for our choice of backbone: ResNet50, EfficientNetB0 and EfficientNetB1. The
results obtained with these backbones are shown in Table \ref%
{table:ablation-cnn}. The two EfficientNet variants achieve a better
performance compared to the ResNet50 backbone, either due to overfitting
(26M parameters for ResNet50 compared to 5.3M and 7.8M for EfficientNetB0
and EfficientNetB1, respectively \cite{tan2019efficientnet}) or a better
learning capacity \cite{tan2019efficientnet}. The best performance is
achieved with the EfficientNetB1 backbone, suggesting that further
improvements in accuracy can be obtained with appropriate deeper models
(e.g., deeper EfficientNet models), at the expense of memory and runtime.
\begin{table}[tbh]
\caption{Ablations of the convolutional backbone of our model, evaluated on
the 7Scenes dataset. We report the average of median position and
orientation errors, across all scenes. The chosen model is highlighted in
bold.}
\label{table:ablation-cnn}\centering%
\begin{tabular}{lcc}
\toprule
\textbf{Backbone} & \textbf{Position} & \textbf{Orientation} \\
& \textbf{[meters]} & \textbf{[degrees]} \\ \midrule
{Resnet50} & 0.19 & 8.60 \\
{\textbf{EfficientNetB0}} & 0.18 & 7.55 \\
{EfficientNetB1} & 0.17 & 7.26 \\\bottomrule
\end{tabular}%
\end{table}

\textbf{Resolution of Activation Maps.} The EfficientNet backbone can be
sampled at different endpoints. As we move along these endpoints, the
receptive field and the depth of each entry grow. Thus, activation maps
sampled at different levels capture different features, which may vary in
the degree to which they are informative for position and orientation
estimation. To evaluate this effect, we train our model by sampling the
position and orientation activation maps, $A_{x}$ and $A_{q}$, at different
endpoints. Specifically, we consider sampling both $A_{x}$ and $A_{q}
$ from the same endpoint or when segregating the sampling from two different
resolutions. Table \ref{tb:ablsation_map_res} shows the results of different
combinations. The best performance is obtained by providing a combination of
coarse and fine activation maps for the position and orientation
transformers, respectively. It follows that the chosen resolutions of $%
14\times 14\times 112$/$28\times 28\times 40$ are a sweet spot in this
parameter space. In particular, for {$\mathbf{A_{x}}$}, the position
estimation, the $14\times 14\times 112$ resolution is optimal as we tested
for both the smaller ($7\times 7\times 320$) and larger ($28\times 28\times
40$) resolutions. As for {$\mathbf{A_{q}}$,} the orientation
estimation, we tested for the smaller ($14\times 14\times 112$) activation
map. We were unable to test for the higher {$\mathbf{A_{q}}$ }%
resolution ($56\times 56\times \cdot $) due to the associated memory
constraints.

\begin{table}[tbh]
\caption{Ablations of activation maps evaluated on the 7Scenes dataset. $%
\mathbf{A_{x}}$ and $\mathbf{A_{q}}$ are sampled at different resolutions
and passed to the respective transformer head. We report the average of
median position and orientation errors across all scenes. The chosen model
is highlighted in bold.}
\label{tb:ablsation_map_res}\centering
\par
\begin{tabular}{ccc}
\toprule
\begin{tabular}{@{}c}
\textbf{Resolution} \\
\textbf{$\mathbf{A_{x}}/\mathbf{A_{q}}$}%
\end{tabular}
&
\begin{tabular}{@{}c}
\textbf{Position} \\
\textbf{[meters]}%
\end{tabular}
&
\begin{tabular}{@{}c}
\textbf{Orientation} \\
\textbf{[degrees]}%
\end{tabular}
\\ \midrule
$28\times28\times40$/$28\times28\times40$ & 0.22 & 7.42 \\
{$\mathbf{14\times14\times112}$/$\mathbf{28\times28\times40}$}& 0.18 & 7.55
\\
$7\times7\times320$/$28\times28\times40$ & 0.19 & 8.32 \\
$14\times14\times112$/$14\times14\times112$ & 0.19 & 7.96 \\ \bottomrule
\end{tabular}%
\end{table}
\textbf{Transformer Architecture} The main hyperparameters of our
Transformer architecture follow the standard choice. Thus, we further
evaluate the sensitivity of our model performance to changing two main
hyperparameters: the number of layers in the encoder and decoder components
and the transformer dimension, $C_{d}$. The results are shown in Table \ref%
{table:ablation-encoder-decoder-layers} and Table \ref%
{table:ablation-transformer-dims}, respectively. All variants considered, in
terms of layer number and transformer dimension, maintain the SOTA position
and orientation accuracy, compared to other APR solutions (Table \ref%
{tb:7scenes_rank}). The accuracy improves with increasing the dimension of
the Transformer, suggesting that larger models may achieve further
improvement in localization accuracy. We note that regardless of the number
of layers (Table \ref{table:ablation-encoder-decoder-layers}), our model
outperforms other solutions (see Table \ref{tb:7scenes_rank}). We
chose the standard 6-layer model for our operations and for comparing it with
our previous multi-scene architecture (to allow for a fair comparison), but
also include a shallower model in runtime and memory analysis.
\begin{table}[tbh]
\caption{Ablations of the number of layers in the Transformer Encoders and Decoders evaluated using the 7Scenes dataset. We report the average median position and orientation errors across all scenes. The chosen model is highlighted in bold.}
\label{table:ablation-encoder-decoder-layers}\centering%
\begin{tabular}{ccc}
\toprule
\textbf{Encoder/Decoder} & \textbf{Position } & \textbf{Orientation} \\
{\textbf{\# Layers}} & {\textbf{[meters]}} & \textbf{[degrees]} \\
\midrule
{2} & {0.19} & {8.21} \\
{4} & {0.18} & {7.62} \\
{\textbf{6}} & {0.18} & {7.55} \\
{8} & {0.18} & {7.47} \\ \bottomrule
\end{tabular}%
\end{table}
\begin{table}[tbh]
\caption{Ablations of the transformer's latent dimension, $C_{d}$, evaluated
on the 7Scenes dataset. We report the average of median position and
orientation errors across all scenes. The chosen model is highlighted in
bold.}
\label{table:ablation-transformer-dims}\centering
\begin{tabular}{ccc}
\toprule
\textbf{Transformer Dimension} & \textbf{Position } & \textbf{Orientation}
\\
{} & {\textbf{[meters]}} & \textbf{[degrees]} \\ \midrule
{64} & {0.19} & {8.32} \\
{128} & {0.18} & {7.32} \\
{\textbf{256}} & {0.18} & {7.55} \\
{512} & {0.18} & {7.21} \\ \bottomrule
\end{tabular}%
\end{table}

\textbf{Number of Clusters} Our coarse-to-fine residual learning approach
assumes the computation of $k$ clusters (and centroids) for each scene, for
orientation and position. To study the effect of this hyperparameter at
small and large scene scales, we train our model with a varying number of
clusters. Table \ref{table:ablation-num-clusters} shows the results of this
experiment for the CambridgeLandmarks and the 7Scenes datasets. We
highlight the model chosen to report our main results by preferring the
position over orientation errors. For mid- and large- scale scenes
(CambridgeLandmarks), $k=4$ yields the best results, with a similar
performance obtained also for other choices (we report the performance of
models trained for 400 epochs). When considering small-scale scenes (7scenes
dataset), the choice of $k$ has a significant effect on position estimation,
with the best performance obtained using a single cluster. The
results show the advantage of partitioning a scene into smaller subareas
when learning to regress positions for mid- or large-scale scenes. When
dealing with small-scale scenes, as in the 7Scenes dataset, this division
does not improve position estimation. When considering orientation
regression, the division improves performance, but, the results are less
consistent and significant. We postulate that this can be improved by
augmenting angle-based clustering with position information, where such an
improvement is a direction for further research.
\begin{table}[tbh]
\caption{Ablations of the number of position and orientation clusters, $%
K_{x} $ and $K_{q}$, respectively. We report the average of the median
position and orientation errors across all scenes for the 7Scenes and
CambridgeLandmarks datasets. The chosen model is highlighted in bold, as we
gave preference to the position error over the orientation error.}
\label{table:ablation-num-clusters}\centering
\begin{tabular}{cccc}
\toprule
&  & \multicolumn{2}{c}{\textbf{Average [m/deg]}} \\
{$K_{x}$} & {$K_{q}$} & \textbf{Cambridge Landmarks} & \textbf{7Scenes} \\
\midrule
{8} & {8} & {1.69/3.06} & {0.25/7.93} \\
{4} & {4} & {\textbf{1.27/3.01}} & {0.24/7.61} \\
{2} & {2} & {1.09/3.16} & {0.20/7.45} \\
1 & 2 & --- & {\textbf{0.18/7.55}} \\ \bottomrule
\end{tabular}%
\end{table}

{\textbf{Scalability.} Runtime and memory footprint are two of the
main advantages of single-scene APRs. However, in order to cover a site with
$N$ scenes, $N$ models need to be stored and selected-from during inference
time. Thus, encoding a site of 1000 scenes with PoseNet models \cite%
{kendall2015posenet} of size $30$MB each, requires $1000\times 30$MB = $3$%
GB. For a given number of scenes, the required memory and run-time of the
proposed scheme are invariant of the network's weights and depend only on
its architecture. To evaluate the run-time and memory footprint, it
is enough to instantiate models for a varying number of scenes and measure
their model size and the runtime of their forward pass (without training).
For our ablation study, we consider hypothetical datasets with an order of
magnitude of $10^1-10^3$ scenes. The results are shown in Table \ref%
{tb:scalability}, where we compare two variants of our model: the
architecture used for the comparative analysis (Section IV-B), with six
layers for each encoder and decoder, and a shallower model with two layers
per encoder/decoder, for which we report similar performance as part of our
ablation study (Section IV-C). The memory footprint of our model remains
relatively constant, where increasing from 4 to 1000 scenes adds only 2MB.
In addition, both variants require less than 80MB before any optimization. The
runtime of the model remains constant in the range of 4-100 scenes and
increases by $\sim \times 1.5$ per 1000 scenes. Assuming a constant scene
selection time, our model is $\times 2-\mathbf{\times }5$ slower than a
single-scene APR. This can be expected due to the run-time complexity of the
MHA operation, which is quadratic with respect to the sequence length
(number of scenes). However, a significant acceleration can be achieved with
recent linear-time MHA formulations \cite{choromanski2020rethinking} and
other optimization methods \cite{vanholder2016efficient}. We demonstrate
this in the next section.}
\begin{table}[th]
\caption{Runtime (in ms) and memory footprint (in MB) as a function of the
number of learned scenes. The results for two instances of our model,
using two and six layers in all encoders and decoders.}
\label{tb:scalability}\centering%
\begin{tabular}{lcccc}
\toprule
\textbf{Num. Scenes} & \multicolumn{2}{c}{\textbf{Runtime [ms]}} &
\multicolumn{2}{c}{\textbf{Memory [MB]}} \\
\multicolumn{1}{c}{\textbf{Num. Layers}} & 2 & 6 & 2 & 6 \\ \midrule
1 & 18.8 & 34.6 & 40.8 & 74.6 \\
4 & 18.8 & 35 & 40.8 & 74.6 \\
7 & 19.2 & 35.2 & 40.8 & 74.6 \\
10 & 19.2 & 35.2 & 40.8 & 74.6 \\
100 & 19.6 & 35.4 & 41.0 & 74.8 \\
500 & 21.0 & 41.0 & 41.8 & 75.6 \\
1000 & 27.0 & 58.6 & 42.8 & 76.7 \\ \bottomrule
\end{tabular}%
\end{table}
\bigskip

{\textbf{Runtime and Memory.} We evaluate and compare the runtime and
model size, and the total storage required, for our method and other
localization schemes, {in Tables \ref{table:model_size} }and \ref%
{table:storage}, respectively. MS-Transformer and the proposed
c2f-MS-Transformer were implemented using the EfficientNet-B0 backbone and
have 6 layer encoders and decoders. The runtime of Duong et al. \cite%
{8491017}, Brachmann et al. \cite{7780735} and DSAC \cite{DSAC} are cited
from \cite{8491017}, while the DSAC* timing is cited from \cite{9394752}.
All timing measurements were made using GTX1080 and Tesla K80 GPUs that are
of similar processing speed. We also applied NVIDIA's TensorRT\footnote{%
https://developer.nvidia.com/tensorrt} (TRT) to optimize our model size and
runtime, using TRT 8.0.3.4 with CUDANN 8.2 and default settings. With this
optimization, our model runs in {6.1}ms ($\times${5.77}
speedup) and has a model size of {48.2}MB ({27}\% reduction),
demonstrating its potential for deployment in real-time applications on
low-end devices. The optimized version is $\times $ {4.9} faster than
the accelerated version of DSAC* \cite{9394752}. The optimization process
was performed with the GTX2080TI GPU. }
\begin{table}[tbh]
\caption{Runtime and model size comparison of our approach and representative localization
methods. The GeForce 1080 and Tesla K80 GPUs are of similar
computing power.}
\label{table:model_size}\centering{\
\begin{tabular}{lccc}
\toprule
\textbf{Network} & \textbf{Time} & \textbf{Model Size} & \textbf{Device} \\
& \textbf{[ms]} & [\textbf{MB]} &  \\ \midrule
{DSAC*\cite{9394752}} & 30 & 30 & 2080TI \\
{DSAC*\cite{9394752}} & 75 & 30 & K80 \\
{DSAC\cite{DSAC}} & 1500 & -- & 1080 \\
{Brachmann\cite{7780735}} & 1000 & -- & 1080 \\
{Duong\cite{8491017}} & 50 & -- & 1080 \\
{PoseNet\cite{kendall2015posenet}} & 7.8 & {26.7} & 1080 \\ \midrule
{Ours (6 layers)} & 35 & 75 & 1080 \\
{Ours (2 layers)} & 19 & 42 & 1080 \\
{\textbf{Ours-TRT (6 layers)}} & \textbf{6.1} & {\textbf{48.2}} & 2080TI \\
\bottomrule
\end{tabular}
}
\end{table}
\begin{table}[tbh]
\caption{The order of magnitude of the storage required by different localization
schemes for 10 and 100 scenes (assuming 1000 images per scene). For RPRs we
assume single image encoding weighs of 5Kb.}
\label{table:storage}\centering{\ {\
\begin{tabular}{lcc}
\toprule
\multirow{2}{*}{\textbf{Method}} & \multicolumn{2}{c}{{\textbf{Scenes\#}}}
\\
& \textbf{10} & \textbf{100} \\ \midrule
SbP \cite{taira2018inloc} & TB & TB \\
SCR\cite{DSAC} & GB & GB \\
SCR \cite{9665967} & MB & GB \\
RPR (\cite{balntas2018relocnet}) & GB & GB \\
Single Scene APR \cite{kendall2015posenet} & MB & GB \\
\textbf{Multi Scene APR (Ours)} & \textbf{MB} & \textbf{MB} \\ \bottomrule
\end{tabular}
} }
\end{table}

\section{Conclusions}

In this work, we propose a novel transformer-based approach for multiscene
absolute pose regression. Using two Transformer Encoders, self-attention is
applied separately to positional and orientational image cues. Thus,
aggregating the activation maps computed by the backbone CNN in a
task-adaptive manner. Our formulation allows agglomerating
non-scene-specific information in the backbone CNN and Transformer Encoders.
Scene-specific information is encoded by a Transformer-Decoder and is
queried per scene. Furthermore, the proposed mixed classification-regression
architecture introduces a novel coarse-to-fine scheme by incorporating scene
clustering. We demonstrate that our approach provides improved
state-of-the-art localization accuracy for both single-scene and multiscene
absolute regression approaches, across outdoor and indoor datasets.

{\small
\bibliographystyle{IEEEtran}
\bibliography{localization}
}

\end{document}